# Stock Type Prediction Model Based on Hierarchical Graph Neural Network

Jianhua Yao
Trine University
Phoenix, USA

Yuxin Dong
Wake Forest University
Winston-Salem, USA

Jiajing Wang
Columbia University
New York, USA

Bingxing Wang
Illinois Institute of Technology
Chicago, USA

Hongye Zheng
Chinese University of Hong Kong
Hong Kong, China

Honglin Qin*
Stevens Institute of Technology
Hoboken, USA

*Abstract*—This paper introduces a novel approach to stock data analysis by employing a Hierarchical Graph Neural Network (HGNN) model that captures multi-level information and relational structures in the stock market. The HGNN model integrates stock relationship data and hierarchical attributes to predict stock types effectively. The paper discusses the construction of a stock industry relationship graph and the extraction of temporal information from historical price sequences. It also highlights the design of a graph convolution operation and a temporal attention aggregator to model the macro market state. The integration of these features results in a comprehensive stock prediction model that addresses the challenges of utilizing stock relationship data and modeling hierarchical attributes in the stock market.

## I. Introduction

The stock market significantly propels the high-quality development and innovation of the economy and holds an important position in the financial system. The application of data mining technology in the stock market has increasingly attracted widespread attention. However, due to the high randomness and non-stationarity of the stock market, most current research only analyzes stock time series data from a single level, which can lead to a loss in the performance of the model. Some studies have pointed out that analyzing stock price fluctuations at a single level can result in a significant loss of valuable information, and multi-level analysis of stock data can significantly enhance the performance of predictive models [1-2].

Previous studies have mostly focused on how to obtain more data sources related to stocks. Many researchers have applied time series analysis techniques to stock trading sequences for stock prediction [3]. Other researchers have attempted to mine financial news, social media, and other textual information to assist in stock prediction [4]. These methods often only pursue an increase in the number of data sources, neglecting the rich hierarchical information contained within the stock sequence data itself. Stock price fluctuations are influenced by multi-level factors, and how to learn a more comprehensive and more interpretable stock feature expression is the core of the work in this paper. In addition, the rich relational structures in the stock market have also been used for predictive tasks recently [5]. For example, stocks belonging to the same or related industries may exhibit similar price trends. Therefore, how to use the stock relationship to introduce the trend characteristics of related stocks to help predict the target stock is also one of the key issues.

In summary, this paper, while introducing the relationship of stocks, fully mines the multi-level information in the stock sequence, mainly facing two challenges:

- How to utilize stock relationship data.
- How to model the hierarchical attributes in the stock market.

In response to the above two issues, this paper designs a Hierarchical Graph Neural Network (HGNN) model based on stock relationships for stock-type prediction tasks. The HGNN can mine stock relationship information and predict stock types by integrating the market status of multiple levels.

Firstly, a stock industry relationship graph is constructed to model the stock relationship information, where each stock is represented as a node, and each edge between stocks represents that the two stocks belong to the same industry.

Secondly, temporal information expression is extracted from the historical price sequence of stocks and serves as the expression of the stock's own state. Then, a graph convolution operation is designed on the stock industry relationship graph to absorb the temporal information of related stocks, generating the industry state expression of the target stock. Subsequently, a temporal attention aggregator is designed to adaptively select and aggregate the trend information of all stocks to model the macro market state. Finally, the features of the three levels are integrated for stock prediction.

## II. Related Work

In the domain of stock market prediction, various approaches have been explored, leveraging advancements in neural networks [6-8], deep learning[9-12], and graph-based models[13-16]. This section reviews the most relevant contributions to these fields, emphasizing the methodologies that align with the proposed Hierarchical Graph Neural Network (HGNN) model.

Stock market prediction has long been a challenging task due to its inherent randomness and non-stationarity. Traditional approaches have often relied on time series analysis and machine learning algorithms to model stock prices. For instance, Sun et al. [17] explored the use of Long Short-Term Memory (LSTM) networks combined with extreme value theory to manage financial risks in high-frequency trading, highlighting the significance of temporal modeling in financial data. Similarly, recurrent neural networks (RNNs) have been adapted to improve predictive accuracy by integrating advanced computational techniques, as demonstrated by Jiang et al. [18], who introduced a Carry-lookahead RNN to enhance processing efficiency.

Beyond temporal models, recent studies have underscored the importance of incorporating hierarchical and relational structures within stock data. Cheng et al. [19] introduced a Graph Neural Network with Contrastive Learning (GNN-CL) for advanced financial fraud detection, emphasizing the utility of graph-based models in capturing complex relationships within financial networks. This approach resonates with the current work, which constructs a stock industry relationship graph to model inter-stock relationships and enhance prediction accuracy. Yiyi Tao et al. [20] proposed the FacT (Factorization Trees) model, which aims to enhance the explainability of latent factor models. In the broader context of machine learning applications, Liu et al. [21] analyzed the impact of external factors, such as epidemics, on predictive models using machine learning algorithms. Although their focus was on a different domain (New York taxi data), their methodology provides insights into the adaptability of machine learning techniques across different data environments. Lastly, the efficiency optimization of large-scale models has been a focal point in recent research. Mei et al. [22] explored deep learning techniques to enhance natural language processing tasks, which, while not directly related to stock market prediction, provides a foundation for optimizing large models like HGNNs, ensuring they can handle the complexity and scale of financial data.

In summary, the proposed HGNN model draws on a rich body of work in time series analysis, graph neural networks, and deep learning optimization. By integrating hierarchical and relational data, the HGNN aims to address the limitations of existing models, providing a more comprehensive and interpretable framework for stock market prediction.

### III. BACKGROUND

On day $t$, a stock collection containing $S_t$ stocks is defined as $S = \{s^1, s^2, \cdots, s^{M_t}\} \cup \{s^{M_t+1}, s^{M_t+2}, \cdots, s^{S_t}\}$, where it includes $M_t$ trading curb stocks $\mathcal{M} = \{s^1, s^2, \cdots, s^{M_t}\} \in S$ within SS. Due to the special nature of trading curb stocks, they will be treated differently subsequently. First, the trading time series data for the past T time steps for each stock in the stock collection S on day $t$ is calculated as $X_t = \{x_{t-T+1}, x_{t-T+2}, \cdots, x_t\} \in \mathbb{R}^{T \times F}$, where FF is the dimension of the input data features (such as opening price, closing price, etc.). In addition, for all trading curb stocks on day t, high-frequency minute-level data is obtained to match the suddenness of the trading curb, and particularly, minute-level data is used to calculate trading curb related technical indicators $D_t = \{d_t^1, d_t^2, \cdots, d_t^{M_t}\}$ (referred to as trading curb related indicators, including moving average, rate of change, turnover rate, amplitude, deviation rate, etc.).

At time step $t$, the stock graph based on industry attributes can be defined as $G_t = \{V_t, E_t\}$, where $V_t$ represents all $S_t$ stock nodes on day $t$; $E_t$ denotes all edges, representing the relational connections within the stock ensemble.

Given a stock relationship graph G, the trading time series data $X_t^{m_t}$, and the trading curb related indicators $D_t^{m_t}$, the stock type prediction task is to learn a mapping Γ to predict the type label $\hat{y}_t$ for trading curb stocks. The label 1 or 0 represents the type of trading curb stock, i.e., whether a stock that has touched the trading curb price will close at the trading curb price (Type I or Type II). The mathematical form is represented as:

$$(X_t^{m_t}, D_t^{m_t}, G) \xrightarrow{\Gamma} \hat{y}_t^{m_t}$$

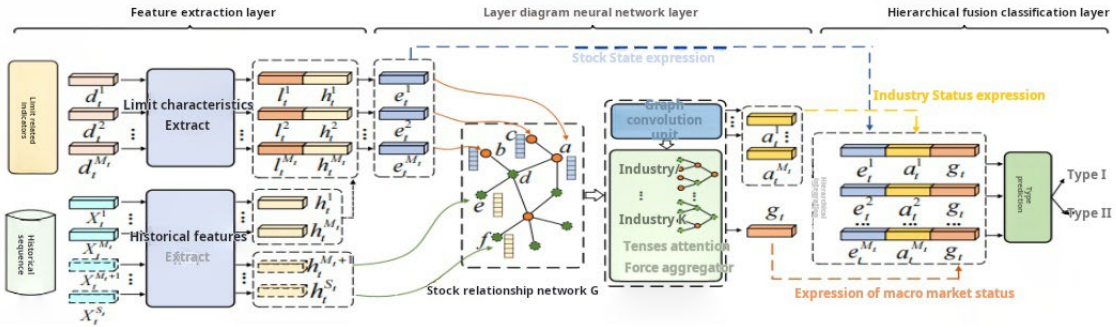

Figure 1. Hierarchical Graph Neural Network (HGNN) Stock Prediction Model Architecture

### IV. METHOD

This paper introduces a Hierarchical Graph Neural Network (HGNN) model proposed for the prediction of stock types. Initially, data on the industry categories of stocks are collected, and a stock industry relationship graph is constructed based on the industrial relationships between stocks. Based on this relationship graph, a hierarchical graph neural network model is designed to expressively capture the stock's own state features, industry state features, and macro market state features at different levels. The stock's own state features can enhance the importance of the stock's own temporal characteristics. The industry state features include the trend information of related stocks, while the macro market state features can indicate the sentiment of the macro market. Figure 1 illustrates the overall architecture of the stock type prediction model HGNN proposed in this paper, which includes three main modules: the Feature Extraction Layer, the Hierarchical Graph Neural Network Layer (HGNN Layer), and the Hierarchical Fusion Classification Layer.

*A. Feature Extraction Layer*

The role of this module is to sequentially extract the time series features and trading curb related features of stocks using price time series data and trading curb technical indicators.

Historical Time Series Feature Extraction. The Historical Feature Extraction unit is utilized to extract historical time series features from the raw price history sequence. The Long Short-Term Memory (LSTM) network is an optimized model of Recurrent Neural Networks (RNNs), which has been proven effective in capturing long-term dependency information in time series data [23-25]. Therefore, this paper employs the LSTM network to extract the historical time series features of stocks. For the historical sequence $X_t^{s_t}$ of stock $s_t$ on day $t$, the LSTM encodes it to obtain the historical time series features, expressed by the following formulas:

$$i_t^{s_t}, r_t^{s_t}, o_t^{s_t} = f_{\theta^i}, f_{\theta^f}, f_{\theta^o}(h_{t-1}^{s_t}; x_t^{s_t})$$

$$u = g(P_u h_{t-1}^{s_t} + Q_u x_t^{s_t} + b_u)$$

$$c_t^{s_t} = r_t^{s_t} \odot c_{t-1}^{s_t} + i_t^{s_t} \odot u$$

$$h_t^{s_t} = o_t^{s_t} \odot g(c_t^{s_t})$$

In this formula, $f$ is a feedforward network with a sigmoid activation function, $\theta$ represents the set of learnable parameters for the gating units; $g$ is the tanh function; $i, r, o$ denote the input gate, forget gate, and output gate, respectively; $P_u \in \mathbb{R}^{U \times K}$ and $Q_u \in \mathbb{R}^{U \times U}$ are the trainable weight matrices, where $U$ is the number of memory cells in the LSTM. After recursively encoding the input historical sequence, the Historical Feature Extraction unit outputs the last hidden feature $h_t^{s_t}$ as the historical time series feature of stock $s_t$.

Trading curb Related Feature Extraction: Due to the existence of the trading curb phenomenon in the stock market, it restricts the abnormal trend of stocks. Considering the special nature of trading curb stocks, this paper uses high-frequency minute-level historical trading data to capture the suddenness of fluctuation and calculates the trading curb related technical indicators $D_t = \{d_t^1, d_t^2, \cdots, d_t^{M_t}\}$ on the high-frequency minute data. To extract the trading curb related features from the trading curb related indicators, the Limit Feature Extraction unit employs a Multi-layer Perceptron (MLP) to implement nonlinear transformation. Given the trading curb related indicators $d_t^{m_t}$ of stock $m_t$, it inputs them into the MLP to extract the trading curb related features $l_t^{m_t}$:

$$l_t^{m_t} = MLP(d_t^{m_t})$$

where $MLP(\cdot)$ is a Multi-layer Perceptron module composed of multiple fully connected layers, with LeakyReLU as the activation function. For the special trading curb stock $m_t$, it transforms the input vector $d_t^{m_t}$ into the trading curb related feature $l_t^{m_t} \in \mathbb{R}^U$, using high-frequency data to better represent the suddenness of limit-ups.

*B. Graph Convolutional Unit*

In this paper, the hidden representations $h_t$ for all $S_t$ stock nodes are initialized using an LSTM network. Due to the particularity of price-limited stocks, there are two types of stock nodes in the stock relationship graph: regular stock nodes $z_t$ and price-limited stock nodes $m_t$. For regular stock nodes $z_t$, the stock's own state expression $e^{z_t}$ is represented as $h^{z_t}$. For price-limited stock nodes $m_t$, the feature expression is related not only to historical time-series features but also to features associated with price limits. Therefore, by interactively fusing historical time-series features with price limit-related features, the stock's own state expression for stock $m_t$ is obtained:

$$e_t^{m_t} = \psi(W_f^T [h_t^{m_t} \oplus l_t^{m_t}] + b_f),$$

where $\psi$ denotes the activation function; $\oplus$ represents the concatenation operation; $W_f$ is a trainable nonlinear transformation matrix. By initializing the node features for all $S_t$ stocks, the stock's own state feature expression (Node View) is obtained.

To simulate the propagation of relational information between stocks and to more comprehensively learn the industry state expressions of stocks, this paper designs a graph convolutional unit that adaptively absorbs the trend information of related stocks. For each stock node $s_t$ in the relationship graph, the temporal feature information of all its neighboring nodes is aggregated in a learnable manner, ultimately obtaining the industry state expression $a_t^{s_t} \in \mathbb{R}^U$ for stock $s_t$ on day $t$. The graph convolution operation can be represented as:

$$a_t^{s_t} = \sum_{j \in N_s} (\pi \cdot e_t^{j_t})/r_{j \cdot s}$$

$$r_{j \cdot s} = \sqrt{\text{degree}(j_t) \cdot \text{degree}(s_t)}$$

where $\text{degree}(s_t)$ is the degree of stock node $s_t$ in the relationship graph; $j_t \in N_{s_t} \cup \{s_t\}$ represents the neighbor nodes of stock node $s_t$ in the stock relationship graph. The features of the related stock nodes are first transformed and encoded by a weight matrix $\pi$. Specifically, the industry state expression (Relation View) for stock $m_t$ is $a^{m_t} \in \{a_t^1, a_t^2, \cdots, a^{M_t}\}$.

Market-Oriented Temporal Attention Aggregator. In the real stock market, the sentiment of the entire market can affect the trend of a single stock. A direct method is to introduce stock indices to guide predictions, but in the information-asymmetric stock market, stock indices can be easily distorted, meaning that the stock indices do not keep pace with the real market. Therefore, this paper proposes a Market-Oriented Temporal Attention Aggregator that introduces temporal attributes and considers the importance of different stocks, dynamically integrating the node features of all stocks in the stock relationship graph to simulate the changing macro market state over time. Thus, the market-oriented temporal awareness weight calculation formula is as follows:

$$w(s_t) = \frac{\exp(\eta_t^{s_t})}{\sum_{j_t} \exp(\eta_t^{j_t})},$$

$$\eta_t^{s_t} = P_a^T \phi(Q_a a_t^{s_t} + b_a)$$

where $\phi$ represents an activation function; $P_a \in \mathbb{R}^V$, $Q_a \in \mathbb{R}^{V \times U}$ are weight matrices that require training.

*C. Hierarchical Fusion Prediction Layer*

This module aims to integrate the stock's own state features, industry state features, and macro market state features from three different levels of information into the final expression of the stock for classifying the type of price-limited stocks.

(1) Hierarchical Fusion: The stock's own state feature $e_t^{m_t}$ is beneficial for enhancing the importance of the target stock's historical time-series features. In addition, the stock's industry state feature $a^{m_t}$ and the macro market state feature $g_t$ respectively reflect the role of the industry relationship graph at different levels, where the industry state feature contains information about the trend fluctuations of related stocks, and the macro market state feature represents the current market sentiment. Considering the hierarchical nature of market states, to learn a more comprehensive and less uncertain stock representation, this paper fuses $e_t^{m_t}$、$a_t^{m_t}$, and $g_t$ together as the final expression $H_t^{m_t} \in \mathbb{R}^{3U}$:

$$H_t^{m_t} = \left[e_t^{m_t} \oplus a_t^{m_t} \oplus g_t\right]^T$$

(2) Stock Type Prediction: Finally, a fully connected layer is implemented as the stock type prediction function, which outputs the type $\hat{y}_t^{m_t}$ of the target price-limited stock $m_t$:

$$\hat{y}_t^{m_t} = Q^T H_t^{m_t} + b$$

where $Q$ is the trainable parameter matrix. Ultimately, the stock type prediction module is used to determine the types of all price-limited stocks.

## V. EXPERIMENT

*A. Experiment Settings*

This section describes the data-related statistical information used in the stock type prediction task, comparative methods, as well as model parameter settings and evaluation metrics.

*1) Dataset*
The data required for stock type prediction mainly includes historical stock sequence data and stock industry classification data. Among them, the historical stock sequence data includes daily frequency historical trading data and high-frequency minute-level trading data for special fluctuation stocks. For our analysis, we utilize the Bloomberg Market Data Feed (B-PIPE), which offers comprehensive access to historical trading data. This includes daily frequency historical trading data encompassing open, high, low, and close prices along with trading volume. Additionally, B-PIPE provides high-frequency minute-level trading data, which is crucial for capturing special fluctuations in stock prices that occur within the trading day. Subsequently, using the stock industry classification data, we construct a stock relationship graph based on industry relationships.

*2) Baselines*
This paper implements and compares traditional machine learning classification methods such as Naive Bayes, Logistic Regression (LR), Support Vector Machine (SVM), and eXtreme Gradient Boosting (XGBoost), with deep learning-based time series prediction methods like Long Short Term Memory (LSTM) and its variant Attentive LSTM (ALSTM), and graph-based methods on stock relationships such as Graph Convolutional Network (GCN) and Graph Attention Network (GAT), with the proposed Hierarchical Graph Neural Network (HGNN).

*B. Experiment Results Analysis*

To verify the effectiveness of the HGNN in type prediction tasks, this section implements various baseline forecasting methods for comparison with the HGNN prediction model, including classical machine learning classification models (Naive Bayes, LR, SVM, XGBoost), time series forecasting models based on recurrent neural networks (LSTM, ALSTM), and graph neural network models based on stock relationships (GCN, GAT). The experimental results are mainly analyzed in detail from three aspects: overall performance, the role of hierarchical information, and the role of stock relationships.

TABLE I EXPERIMENT RESULTS

| Method | SSE | |
|---|---|---|
| | Acc(%) | F1(%) |
| Naïve Bayes | 46.21±1.31 | 44.37±1.60 |
| LR | 58.59±1.03 | 59.43±0.95 |
| SVM | 58.04±2.11 | 58.90±2.05 |
| XGBoost | 60.59±0.80 | 61.11±0.69 |
| LSTM | 59.74±2.43 | 58.70±1.62 |
| ALSTM | 57.45±2.00 | 57.42±1.38 |
| GCN | 60.54±1.91 | 59.54±1.43 |
| GAT | 61.56±1.99 | 60.55±1.56 |
| HGNN_M | 63.33±1.15 | 63.36±1.11 |
| HGNN_I | 63.74±1.64 | 63.97±1.05 |

As shown in Table I, the experimental results demonstrate that HGNN models outperform other baseline methods in type prediction tasks. HGNN_I shows the best performance with an SSE of 63.74±1.64 and an F1 score of 63.97±1.05%. Classical ML models like Naive Bayes and LR have lower accuracy, while XGBoost performs better. Recurrent neural networks (LSTM, ALSTM) and graph-based models (GCN, GAT) offer moderate results, but HGNN, leveraging hierarchical and stock relationship information, achieves superior accuracy and F1 scores.

## VI. CONCLUSIONS

This paper proposes a Hierarchical Graph Neural Network (HGNN) stock prediction model based on the relationship graph. HGNN constructs a stock industry relationship graph through industry relationships and designs a hierarchical graph neural network method to extract multi-level market state information, including the stock's own state information, the state information of the industry it belongs to, and the macro market state information. The hierarchical structure design of HGNN can explicitly model multi-level market states and learn more comprehensive and more interpretable stock feature expressions from stock data. Extensive experiments on real stock market datasets demonstrate that the proposed stock prediction model HGNN has the best predictive performance, and its multi-level analysis of stock data based on the relationship graph has a clear advantage in the stock prediction task.